\documentclass[letterpaper, 10 pt, conference]{ieeeconf}  

\IEEEoverridecommandlockouts                              

\def\arxiv{1}

\usepackage{graphics} 
\usepackage{graphicx}
\usepackage{epsfig} 
\usepackage{mathptmx} 
\usepackage{times} 
\usepackage{amsmath} 
\usepackage{amssymb}  
\usepackage{tablefootnote}
\usepackage{setspace}

\usepackage{multirow}
\usepackage{makecell}
\usepackage{booktabs}
\usepackage[T1]{fontenc}
\usepackage{newtxtext, newtxmath}
\usepackage{cite}

\usepackage{caption}
\usepackage{subcaption}
\usepackage[table]{xcolor}
\definecolor{lightblue}{RGB}{173, 216, 230} 
\usepackage{graphicx}

\let\labelindent\relax

\usepackage{enumitem}
\usepackage{hyperref}

\newcommand{\methodname}{FLaRe}

\title{\LARGE \bf
\methodname{}: Achieving Masterful and Adaptive Robot Policies with Large-Scale Reinforcement Learning Fine-Tuning
}

\author{Jiaheng Hu$^{1,2}$,  Rose Hendrix$^{1}$, Ali Farhadi$^{1,3}$, Aniruddha Kembhavi$^{1,3}$, \\ Roberto Mart{\'\i}n-Mart{\'\i}n$^{2}$, Peter Stone$^{2,4}$, Kuo-Hao Zeng$^{1, \dagger}$, and Kiana Ehsani$^{1, \dagger}$
\thanks{$^{1}$ Allen Institute for Aritifical Intelligence (Ai2) 
}
\thanks{
$^{2}$ University of Texas, Austin\:
$^{3}$ University of Washington\:
$^{4}$ Sony AI   }
\thanks{$^{\dagger}$ Equal Supervision.}
}

\begin{document}

\maketitle
\thispagestyle{empty}
\pagestyle{empty}

\begin{abstract}

In recent years, the Robotics field has initiated several efforts toward building generalist robot policies through large-scale multi-task Behavior Cloning. 
However, direct deployments of these policies have led to unsatisfactory performance, where the policy struggles with unseen states and tasks.
How can we break through the performance plateau of these models and elevate their capabilities to new heights?
In this paper, we propose \methodname{}, a large-scale Reinforcement Learning fine-tuning framework that integrates robust pre-trained representations, large-scale training, and gradient stabilization techniques. Our method aligns pre-trained policies towards task completion, achieving state-of-the-art (SoTA) performance both on previously demonstrated and on entirely novel tasks and embodiments. Specifically, on a set of long-horizon mobile manipulation tasks, \methodname{} achieves an average success rate of 79.5\% in unseen environments, with absolute improvements of +23.6\% in simulation and +30.7\% on real robots over prior SoTA methods. By utilizing only sparse rewards, our approach can enable generalizing to new capabilities beyond the pretraining data with minimal human effort. Moreover, we demonstrate rapid adaptation to new embodiments and behaviors with less than a day of fine-tuning. 
Videos can be found on the project website at \url{robot-flare.github.io}

\end{abstract}


\section{INTRODUCTION}

Foundation models in computer vision and natural language processing have recently achieved groundbreaking successes. Large-scale transformer models, such as GPT~\cite{achiam2023gpt} and SAM~\cite{kirillov2023segment}, have demonstrated the ability to perform an extensive range of tasks. Inspired by these advances, the robotics community has set its sights on training high-capacity, multi-task transformers for robotic applications.

One of the prominent methods in this pursuit is large-scale behavior cloning (BC) \cite{schaal2003computational}, which leverages large datasets of real-world and simulated demonstrations (e.g., RT-1~\cite{brohan2022rt}, RT-2~\cite{brohan2023rt}, RT-X~\cite{o2024open}, and SPOC~\cite{ehsani2024spoc}) to train high-capacity policies that can perform many different tasks. While BC policies have shown promise, they remain fundamentally limited when directly deployed in the real world: models are constrained to the states observed during training, making it difficult to generalize beyond expert trajectories. Consequently, these policies often struggle when faced with unfamiliar states, and fail to recover from errors effectively.

On the other hand, reinforcement learning (RL) \cite{sutton2018reinforcement} offers a complementary approach that directly optimizes the performance of the robot through trial-and-error learning, and RL algorithms have achieved many successes when a well-defined reward function is available~\cite{tang2024deep, zeng2024poliformer, zhu2020ingredients}. 
However, many RL algorithms are notoriously sample inefficient, requiring extensive training time. As the task horizon increases and the action space grows, RL policies struggle to get off the ground due to the large search space. Moreover, RL's reliance on hand-crafted reward functions severely limits its scalability.

\begin{figure}[t]
\centering
\includegraphics[width=\linewidth]{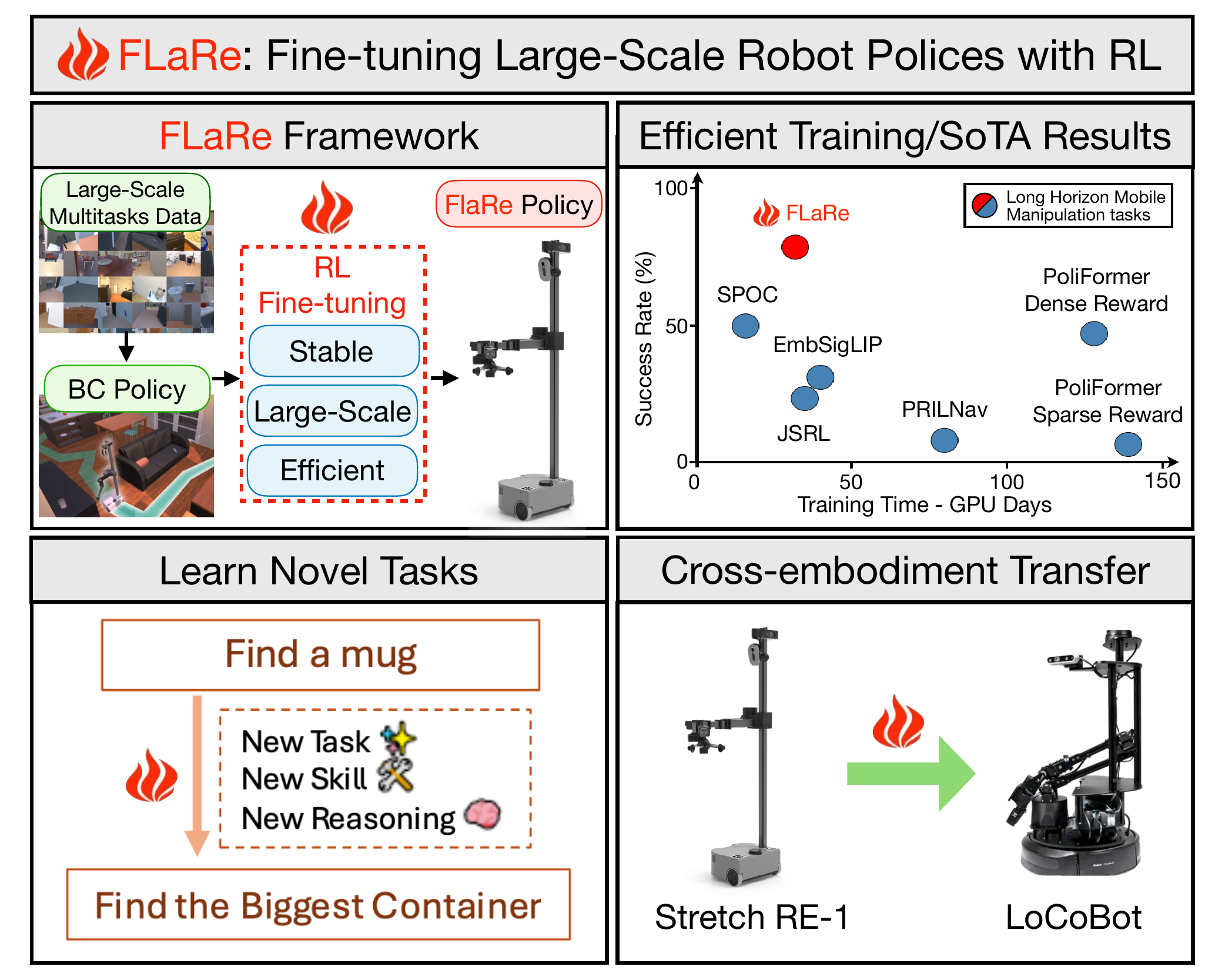}
\caption{\methodname{} is a simple but effective approach for large-scale fine-tuning of robotic policies. \methodname{} achieves SoTA performance on simulation (+23.6\%) and real-world (+30.7\%) benchmarks, can generalize to unseen tasks, and adapts to new behaviors and embodiments. 
}
\label{fig:pull}
\vspace{-2em}
\end{figure}

Although insufficient for direct deployment, the policies trained through large-scale multi-task Behavior Cloning already possess extremely valuable features and behavior priors. 
How can we break through the performance plateau of these models and elevate their capabilities to new heights?
Our key insight is that, through RL, we can align the behavior of these policies towards true objectives such as task completion (instead of the BC objective), thereby achieving masterful performance not only on tasks seen during BC training, but also on novel tasks and embodiments never seen by the pre-trained policy.

While attempts have been made to fine-tune BC policies with RL \cite{uchendu2023jump,ramrakhya2023pirlnav,rajeswaran2017learning,agarwal2022reincarnating}, these works are only verified with small-scale networks and in single-task domains. Empirically, we find these methods ineffective as the capacity of the pre-trained policy scales up, where the abrupt shift from BC to RL often results in destructive gradient updates, leading to oscillations or even collapse during RL training.



In \methodname{}, we introduce an effective, scalable, and robust solution for fine-tuning large-scale robot policies with Reinforcement Learning. 
Illustrated in Fig.~\ref{fig:pull} top-left, \methodname{} starts from a \textbf{multi-task} robotics policy, and fine-tunes it with \textbf{large-scale} RL through extensive use of simulation. To ensure the RL fine-tuning is \textbf{stable}, \methodname{} introduces a set of simple yet highly effective techniques, detailed in Sec.~\ref{ss:ft}, that drastically improve performance and reduce training time compared to previous methods.
\methodname{} achieves SoTA performance on household mobile manipulation tasks. In established simulation benchmark~\cite{ehsani2024spoc}, it achieves an average $79.5\%$ success rate, $+23.6\%$ absolute improvements over the best baseline. In the real world, \methodname{} achieves excellent results ($80.7\%$ SR on average), outperforming the best prior work by $+30.7\%$.
Furthermore, \methodname{} offers several key advantages: 
\begin{enumerate}[itemsep=0pt]
    \item \methodname{} enables efficient training with a 15x reduction in training time compared to the previous SoTA method, using a simple sparse reward without the need for handcrafted reward functions (Fig~\ref{fig:pull} top-right). 
    \item \methodname{} allows for generalization beyond the tasks seen during BC. Even for new tasks without expert trajectories or shaped rewards, \methodname{} can be fine-tuned to achieve state-of-the-art performance (Fig~\ref{fig:pull} bottom-left).
    \item \methodname{} facilitates rapid adaptation to new embodiments and behaviors, significantly enhancing the base model's flexibility and applicability (Fig~\ref{fig:pull} bottom-right).
\end{enumerate}

We find that \methodname{} marks a promising achievement towards developing highly generalizable robotic systems that can handle a wide range of tasks in diverse environments.

\section{Related Work}

\subsection{Foundation model for robotics}

Following the successes of foundation models in vision\cite{kirillov2023segment} and language\cite{achiam2023gpt}, there has been a recent trend towards training robotics-specific foundation models \cite{firoozi2023foundation,hu2023toward}.
While these models focus on different robot applications, such as manipulation (e.g. RT-1 \cite{brohan2022rt}, RT-2 \cite{brohan2023rt}, RT-X\cite{o2024open}, Octo \cite{team2024octo}, RoboCat \cite{bousmalis2023robocat}, OpenVLA \cite{kim2024openvla}),
navigation (e.g. ViNT \cite{shah2023vint}), and mobile manipulation (e.g. SPOC\cite{ehsani2024spoc}), they share a similar recipe of training a high-capacity transformer model through multi-task behavior cloning \cite{schaal2003computational}. 
As a result, they generate the same end-product: a multi-task transformer policy, which \methodname{} can use as a base model for fine-tuning.

\subsection{RL training and fine-tuning of robotics models}

While RL has achieved many successes in robotics\cite{tang2024deep}, directly applying RL from scratch often requires extensive reward engineering and long training time \cite{akkaya2019solving,hu2023causal,zeng2024poliformer,zhu2020ingredients}.
Hence, previous works have extensively explored leveraging pretrained models to facilitate RL \cite{taiga2023investigating,khetarpal2022towards,taylor2009transfer,nair2020awac,agarwal2022reincarnating,hester2018deep,gupta2019relay,julian2020never,vecerik2017leveraging,kober2010imitation,lu2021aw,rajeswaran2017learning,baker2022video,ramrakhya2023pirlnav,zhu2023transfer,wołczyk2024finetuningreinforcementlearningmodels,uchendu2023jump, hu2023imitation,xing2024bootstrapping}.

However, many of these approaches focus on fine-tuning models that have been pre-trained using either online RL \cite{taiga2023investigating,khetarpal2022towards,taylor2009transfer} or offline RL \cite{nakamoto2024cal,kostrikov2021offline,lee2022offline,kumar2022pre}, which limits their applicability. This makes them unsuitable for fine-tuning most existing robotics foundation models, which are typically trained using Behavior Cloning.
Many previous works also require access to the entire offline dataset during fine-tuning\cite{nair2020awac,agarwal2022reincarnating,hester2018deep,gupta2019relay,julian2020never,vecerik2017leveraging,kober2010imitation,lu2021aw,rajeswaran2017learning}, which may be feasible for small-scale data and low-dimensional observations but is unlikely to be computationally feasible for large-scale data and image observations, as also noted by Ramrakhya \textit{et al}~\cite{ramrakhya2023pirlnav}.

In addition, the techniques proposed in many of these works are only evaluated on simple domains, with low-dimensional state spaces~\cite{rajeswaran2017learning,gupta2019relay,nair2020awac}, small-scale network architecture (e.g. MLP)\cite{rajeswaran2017learning,uchendu2023jump,agarwal2022reincarnating}, single-task pretraining and fine-tuning \cite{zhu2018reinforcement,kober2010imitation}, and often no real robot experiments \cite{agarwal2022reincarnating,baker2022video,gupta2019relay,nair2020awac}. PIRLNav \cite{ramrakhya2023pirlnav} and JSRL \cite{uchendu2023jump} are two works that are closest to our setting, where only a pretrained policy is required for the fine-tuning phase. However, both of them focus on single-task setting with small-scale networks and no real robot experiments. 
In contrast, \methodname{} explores fine-tuning from large robotics models, where both scalability and applicability to real robots are of critical concern. 





\section{Problem Formulation}
\label{sec:pf}

We consider each robotics task $T \in \mathcal{T}$ as a language-conditioned Partially Observable Markov Decision Process ($\mathcal{S}$, $\mathcal{A}$, $\mathcal{P}$, $R$,  $\mathcal{O}$, $\mathcal{L}$, $P(s_0)$, $\gamma$), where $\mathcal{S}$ is a state space, $\mathcal{A}$ is an action space, $\mathcal{O}$ is an observation space, $\mathcal{P}$ is a Markovian transition model, $\mathcal{L}$ is a set of natural language instruction, $\gamma$ is a discount factor, $P(s_0)$ is the initial state distribution, and $R$ is a \textbf{sparse} reward function that takes in a natural language instruction $l \in \mathcal{L}$ and a state $s \in \mathcal{S}$ and outputs a binary value indicating whether a given instruction is successfully completed. 
For the purpose of this paper, we assume that all tasks have the same action space (the actuators of the robot) and observation space (the robot's sensors).
Each task $T \in \mathcal{T}$ defines a set of natural language instructions $\mathcal{L}_T$ (e.g., for the task of Object Navigation, potential instructions can be ``go to an apple'', ``find a houseplant'', and more). At the start of every episode, an instruction $l_T \in \mathcal{L}_T$ and an initial state $s_0 \sim P(s_0)$ will be sampled. Every time a specific task $T \in \mathcal{T}$ is given, our goal is to train a policy $\pi^T_\theta$ that maximizes the expected return (i.e. success rate) $\mathbb{E}_{\mathcal{L}_T, \pi} \sum_t R(s_t, l)$ for the given task over the possible language instructions $\mathcal{L}_T$.

\begin{figure*}[t]
\centering
\includegraphics[width=0.9\textwidth]{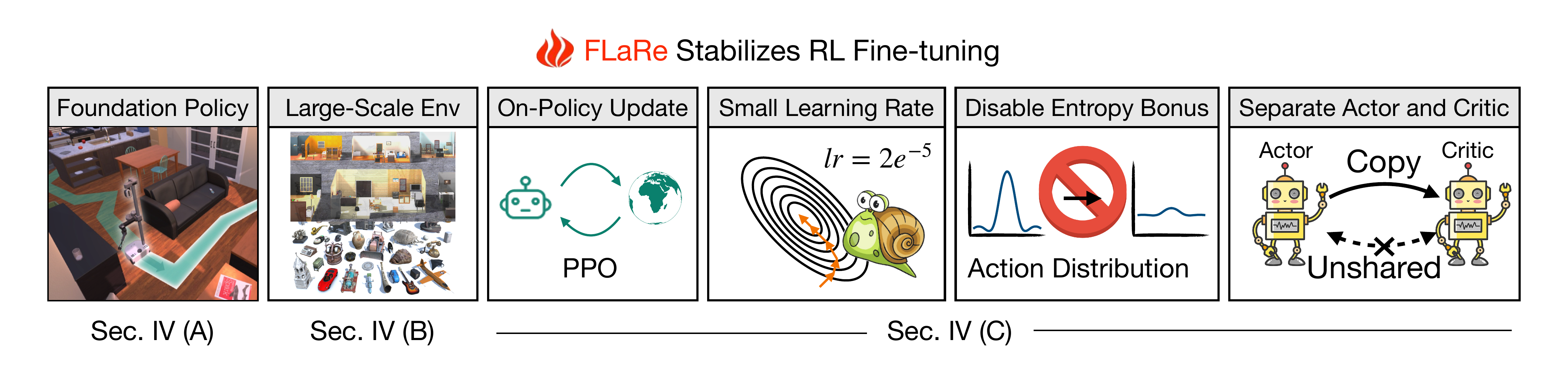}
\caption{
\methodname{} introduces a series of design choices that help stabilize the RL training process, including 1) fine-tuning from a multi-task robotics policy, 2) large-scale fine-tuning in simulation, 3) using an on-policy algorithm as opposed to off-policy methods, 4) utilizing smaller learning rate than when performing RL from scratch, 5) disabling the entropy bonus objective that can potentially distort the policy at the start of the training, and 6) separating the actor and the critic network, so that the critic update will not influence the policy prediction.}
\label{fig:sysdiag2}
\vspace{-1em}
\end{figure*}

\section{Method}

Considerable effort has been devoted to optimizing performance on robotics tasks via training high-capacity models $\pi_\theta$ with large-scale, multi-task imitation learning\cite{ehsani2024spoc,brohan2022rt,brohan2023rt,o2024open}.
In practice, these efforts lead to unsatisfactory performance due to compounding errors \cite{ross2011reduction}, where small action prediction error leads to state distribution drift.
Furthermore, for novel tasks and scenarios where no demonstration data is available, these models have shown limited generalization capabilities, likely due to the limited task coverage of the training data.

\methodname{} addresses both problems by fine-tuning the pre-trained model $\pi_\theta$ with RL for each given task $T \in \mathcal{T}$.
The key idea of \methodname{} is to achieve stable and effective RL fine-tuning through a series of design choices, including 1) utilize a large-scale multi-task model as the base model, 2) achieve large-scale fine-tuning through extensive use of simulations, and 3) a series of algorithmic design to stabilize the RL fine-tuning. 
Together, these design choices enable \methodname{} to effectively learn from \textbf{sparse} reward and achieve good performances.
We elaborate in detail on each of these decisions in the following sections (Fig.~\ref{fig:sysdiag2}).


\begin{figure}[t]
\centering
\includegraphics[width=0.48\textwidth]{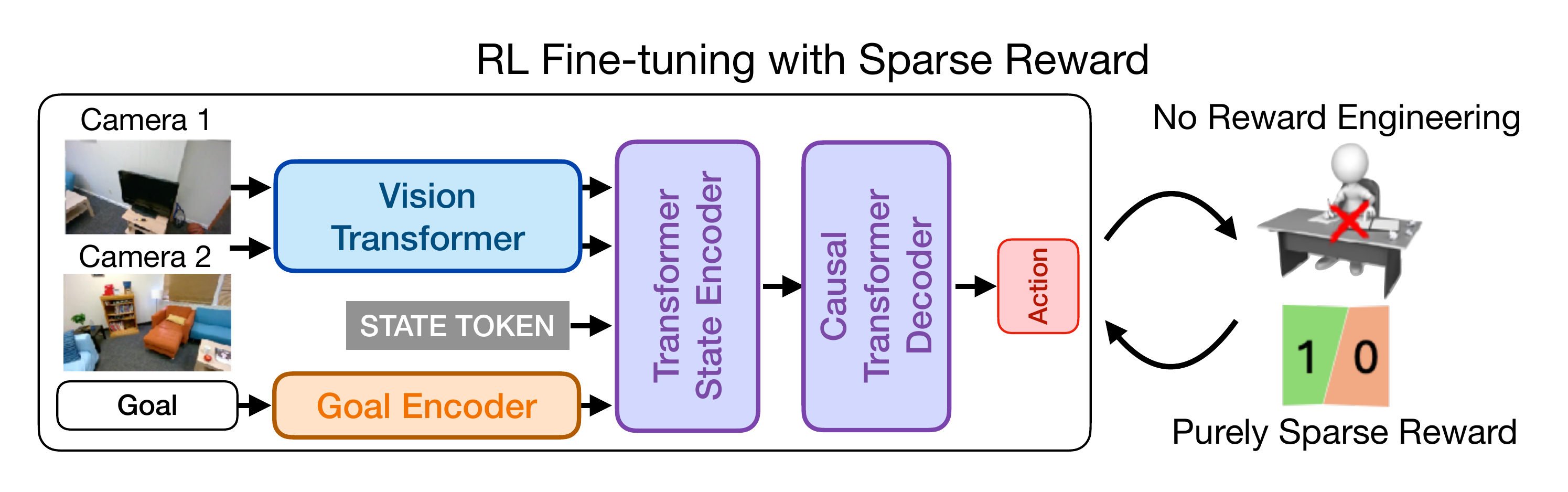}
\caption{
\methodname{} can efficiently fine-tune large transformer policies through large-scale Reinforcement Learning, using a sparse reward function that requires minimal human effort.}
\label{fig:arch}
\vspace{-1em}
\end{figure}

\subsection{Fine-tune from a multi-task robotics model}

The first key design choice of \methodname{} is to start from a \textbf{multi-task} pre-trained large model (i.e. a foundational robotics model). Compared to fine-tuning from a single-task, small-scale network (as is often the case in previous works \cite{uchendu2023jump,rajeswaran2017learning,ramrakhya2023pirlnav,zhu2018reinforcement}), starting from a robotics foundation model brings three key benefits. First, models pre-trained on diverse tasks can master more robust representations and more versatile behavior priors, which will benefit the fine-tuning process. Second, the highly capable network architecture (e.g. large transformer models) that comes with these foundational robotics models brings good inductive bias that can facilitate generalization, which is crucial to fine-tuning. Most importantly, the multi-task capability of these models allows us to reuse the same model for fine-tuning for many different tasks. In fact, as we will show in the experiments in Sec.~\ref{ss:ood}, we can even fine-tune for tasks and embodiments that have never been seen by the pre-trained policy and still achieve good performance.

While our method can in principle work on any foundational robotics model, in this specific work, we focus on fine-tuning the SPOC model (Fig.~\ref{fig:arch})~\cite{ehsani2024spoc} --- a multi-task transformer model for mobile manipulation tasks, trained on large-scale shortest path expert
trajectories collected in Objaverse-Populated ProcTHOR houses\cite{kolve2017ai2,deitke2023objaverse,deitke2022️}. 
Please find more details regarding the SPOC model in App.~\ref{app:spoc_model}.

\begin{table*}[t]
\centering
\footnotesize
\caption{Success and Episode-length weighted Success (SEL) against baseline methods on the CHORES\cite{ehsani2024spoc} benchmark. Baselines with privileged information are \colorbox{lightblue}{\textcolor{black}{\emph{marked in blue}}}. \methodname{} significantly outperforms the previous SoTA methods. }
\begin{tabular}{l||c|c|c|c|c|>{\columncolor{lightblue}}c|>{\columncolor{lightblue}}c}  
\toprule
\multirow{2}{*}{Success (SEL)} 
    & \multicolumn{3}{c|}{\textbf{IL+RL: Sparse Reward}} 
    & \multicolumn{1}{c|}{\textbf{IL Only}} 
    & \multicolumn{3}{c}{\textbf{RL Only}} \\ 
\cmidrule(lr){2-4} \cmidrule(lr){5-5} \cmidrule(lr){6-8}
    & \methodname{} (Ours) & PIRLNav & JSRL & SPOC & Poliformer - Sparse & Poliformer - Dense & EmbSigLIP - Dense \\ 
\midrule

ObjectNav 
    & 85.0 \textbf{(67.6)}
    & 20.0 (7.0)  
    & 21.0 (15.6) 
    & 55.0 (42.2) 
    & 14.5 (10.4)
    & \textbf{85.5} (61.2) 
    & 36.5 (24.5) \\

Fetch  
    & \textbf{66.9 (54.7)} 
    & 0.0 (0.0)
    & 2.9 (2.8) 
    & 14.0 (10.5)
    & 0.0 (0.0)
    & 0.0 (0.0) 
    & 0.0 (0.0)\\

PickUp 
    & \textbf{91.8 (90.4)} 
    & 0.0 (0.0) 
    & 50.9 (47.7) 
    & 90.1 (86.9) 
    & 0.0 (0.0) 
    & 90.1 (88.7) 
    & 71.9 (52.9) \\

RoomVisit 
    & \textbf{70.4 (67.1)} 
    & 12.5 (11.0)  
    &  19.0 (18.6) 
    & 40.5 (35.7) 
    & 12.5 (12.5)
    & 12.5 (10.9)
    & 16.5 (11.9)\\

\bottomrule
\end{tabular}
\label{tab:indist}
\vspace{-1em}
\end{table*}

\subsection{Large-scale fine-tuning in simulation}
\label{ss:sim}
The second key design choice of \methodname{} is to perform large-scale fine-tuning through extensive use of simulation. Recent advancements in robotics and embodied AI have given us a set of tools for simulating robotics tasks\cite{kolve2017ai2, li2023behavior, jin2023mini, puig2023habitat, zhu2020robosuite, xiang2020sapien}. In this work, we utilize AI2THOR \cite{kolve2017ai2} to perform large-scale simulated fine-tuning with diverse objects and scenes, which includes 150k procedurally generated PROCTHOR houses\cite{deitke2022️} and 800K+ annotated 3D objects\cite{deitke2023objaverse}.

When using simulation in robotics, addressing the sim-to-real gap \cite{zhao2020sim} becomes a critical challenge. In \methodname{}, similar to Ehsani \textit{et al.}\cite{ehsani2024spoc}, we employed two techniques to facilitate sim-to-real transfer. First, we perform extensive domain randomization, including color augmentation, applying random crops, and posterizing the images. Second, we extract visual features through DinoV2~\cite{oquab2023dinov2}, a pre-trained foundational vision model, which captures useful features that can generalize across simulation and the real world. 

To ensure large-scale training of the transformer policy and value networks, we utilize the KV-cache technique\cite{pope2023efficiently} to reduce the computational costs during network inference, similar to Zeng \textit{et al.}\cite{zeng2024poliformer}. 
The KV-cache technique caches and reuses the keys and values of earlier observations within an episode. This reduces the inference complexity of the transformer network from quadratic to linear, which is crucial for affordable large-scale RL fine-tuning.


\subsection{Stabilize RL fine-tuning}
\label{ss:ft}
Finally, we introduce a set of simple but very critical algorithmic choices to ensure the stability of RL fine-tuning. 
While these techniques are relatively simple, as we will show in the ablation studies in Sec.~\ref{ss:abla}, each choice is very important to ensure stable training and to obtain good performances. 



\textbf{Using On-policy Algorithms.} 
Off-policy RL methods \cite{mnih2015human, haarnoja2018soft} can utilize off-policy data during training, and thus bring the promise of sample-efficient RL. However, compared to on-policy methods, off-policy RL is often less stable and more sensitive to hyperparameters, both in theory and in practice, due to problems associated with the “deadly triad” \cite{sutton2018reinforcement}. 
In this work, since we perform fine-tuning entirely in simulation, we are less constrained by the sample efficiency of our RL algorithms, and therefore choose to use on-policy algorithms for stable fine-tuning. Specifically, we use PPO \cite{schulman2017ppo}, a state-of-the-art on-policy policy gradient method.


\textbf{Taking Small Update Steps.}
When setting the learning rate for RL, it is common practice to reuse a learning rate that has previously achieved success in the same/similar domains. However, what we found in \methodname{} is that fine-tuning from an existing policy requires significantly lower learning rates than when starting from scratch. For example, in the object navigation task, the previous state-of-the-art result is achieved with PPO from scratch using a learning rate of $2e-4$. In \methodname{}, when fine-tuning on the exact same task, we have to reduce the learning rate by an order of magnitude to achieve stable learning. It is important to notice that we do not perform additional LR tuning in \methodname{}  - the same learning rate is used for all experiments and tasks.

\textbf{Disabling Entropy Bonus.}
The PPO objective \cite{schulman2017ppo} contains an entropy bonus, which promotes the entropy of the action distribution predicted by the policy network to ensure sufficient exploration. However, we found that when fine-tuning from a pre-trained policy network, this additional entropy term can quickly distort the policy gradient update at the start of the training, leading to unlearning of the pre-trained policy. Hence, we remove this entropy bonus term from our PPO update in \methodname{}.

\textbf{Disabling Feature Sharing.}
When applying RL to high-dimensional observations such as images, a standard practice is to have a shared feature extractor between the actor and the critic network, which can facilitate the learning of useful features. However, we found that feature sharing during RL fine-tuning can actually hurt the performance since the gradient from the critic loss will change the pre-trained features and lead to the deterioration of the action prediction.
Furthermore, during RL fine-tuning, since the pre-trained foundation model should already capture good representations, there is no need for the actor and the critic network to share the same feature extractor. Therefore, in \methodname{}, we initialize the policy and the critic network as independent networks, both using the weight and architecture of the pre-trained transformer policy. The policy head of the critic network is replaced by a randomly initialized values.

We found that all four training components are important, and in Section~\ref{ss:abla}, we show that removing any one of them results in training collapse.

\begin{figure*}[t]
\centering
    \begin{subfigure}[b]{0.24\textwidth}
        \centering
        \ifx\arxiv\undefined
        \includegraphics[width=\linewidth]{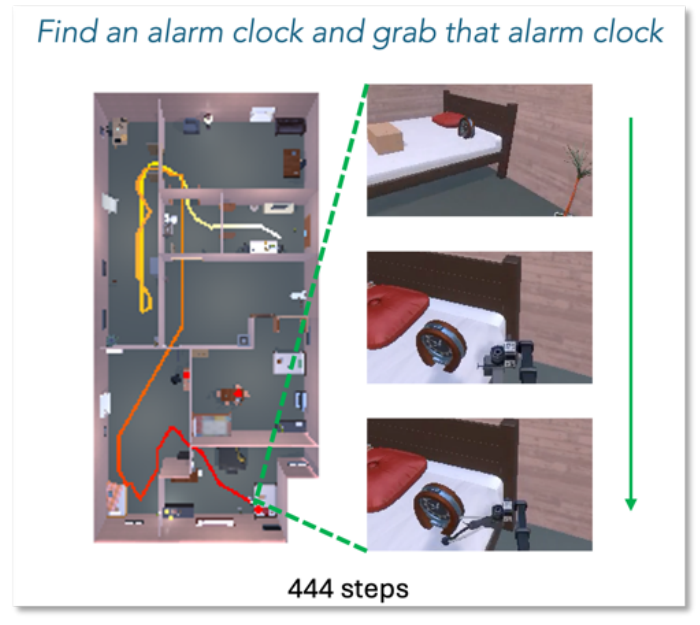}
        \else
        \includegraphics[width=\linewidth]{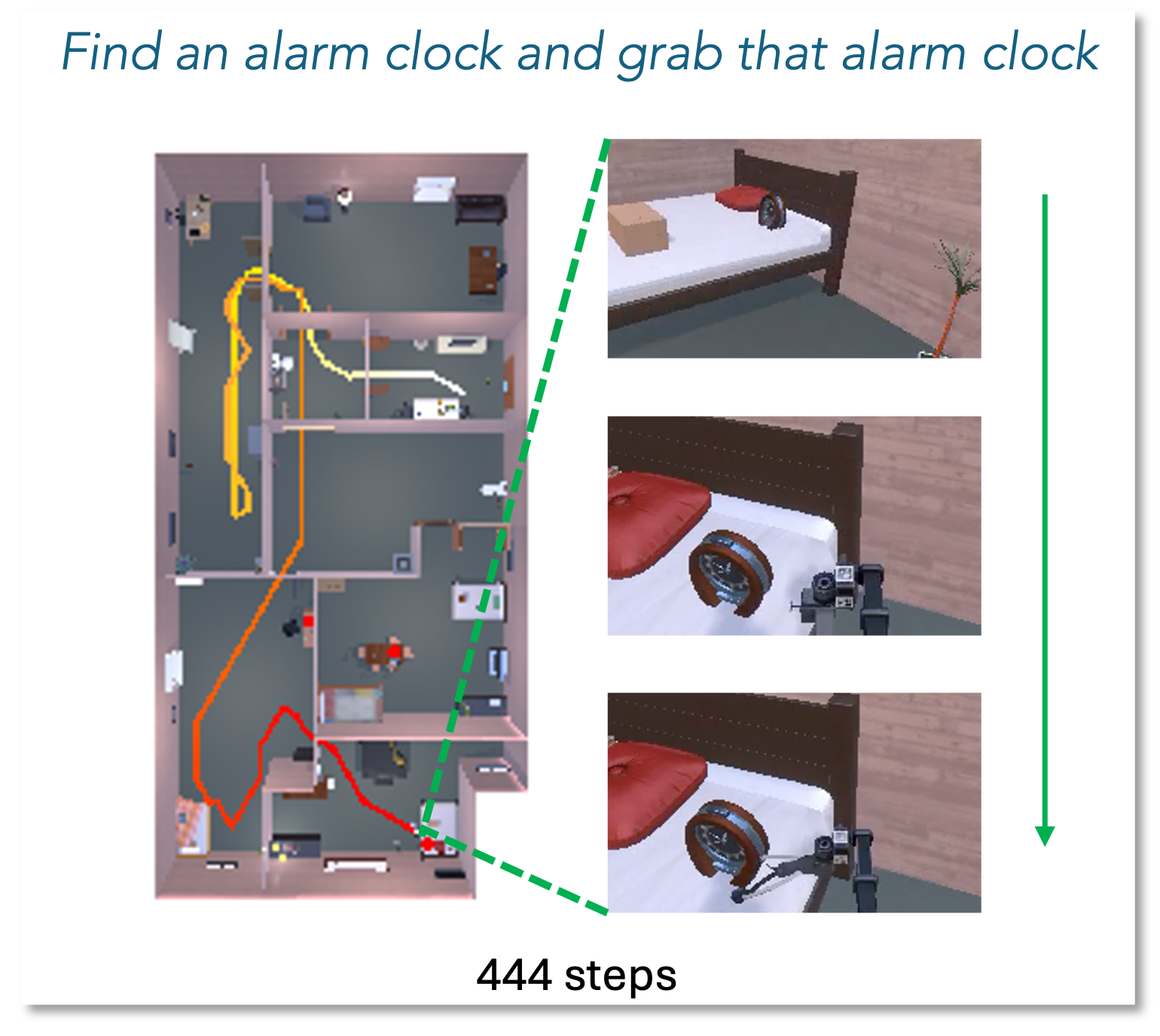}
        \fi
        \caption{Fetch Task}
    \end{subfigure}
    \hfill
    \begin{subfigure}[b]{0.25\textwidth}
                \ifx\arxiv\undefined
        \includegraphics[width=\linewidth]{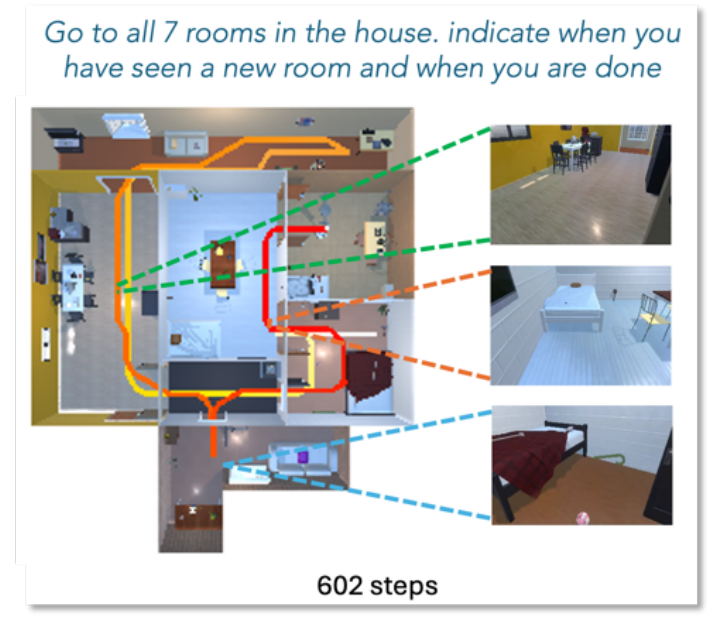}
        \else
        \includegraphics[width=\linewidth]{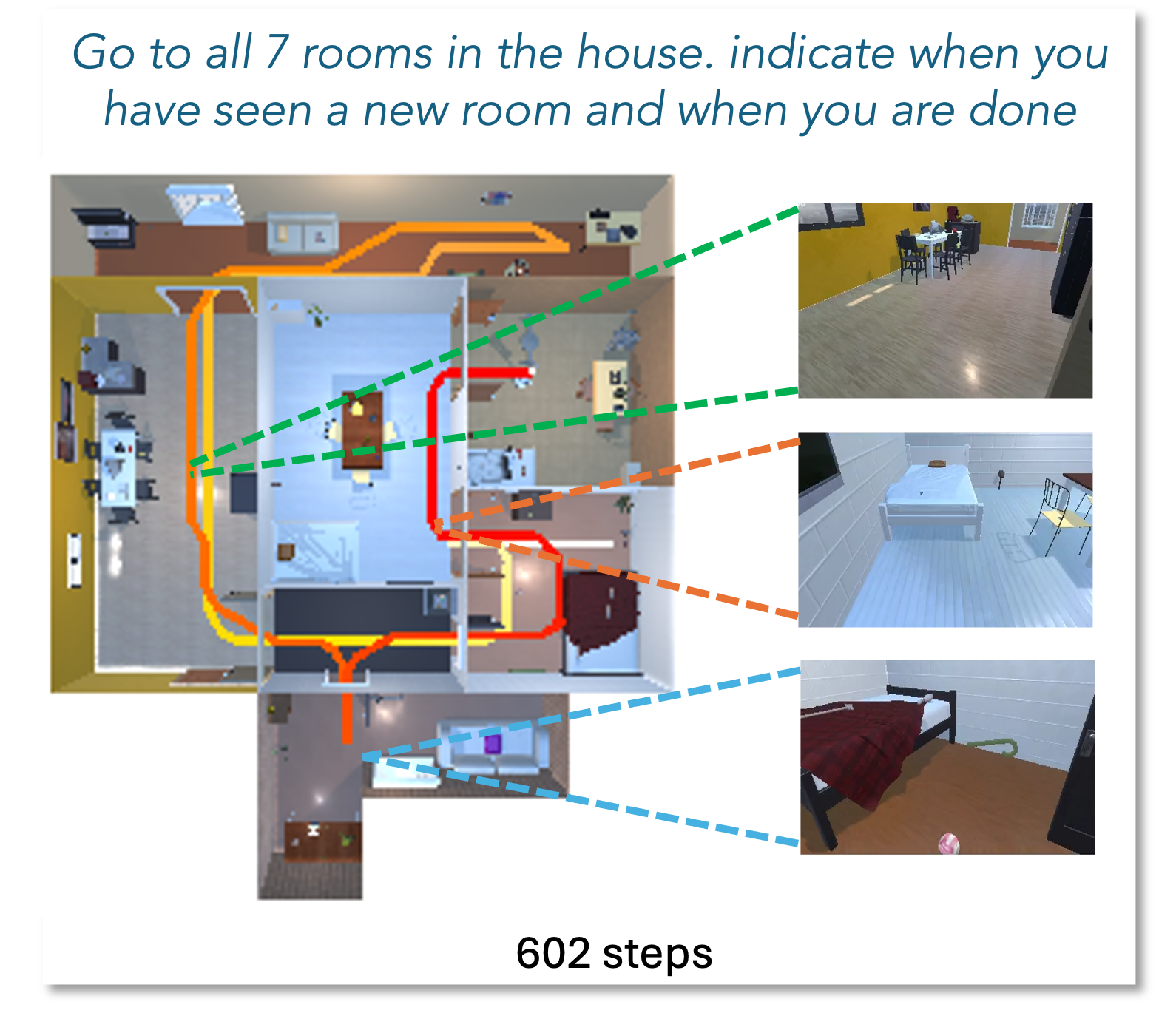}
        \fi
        \caption{RoomVisit Task}
    \end{subfigure}
    \hfill
    \begin{subfigure}[b]{0.24\textwidth}
        \centering
                \ifx\arxiv\undefined
        \includegraphics[width=\linewidth]{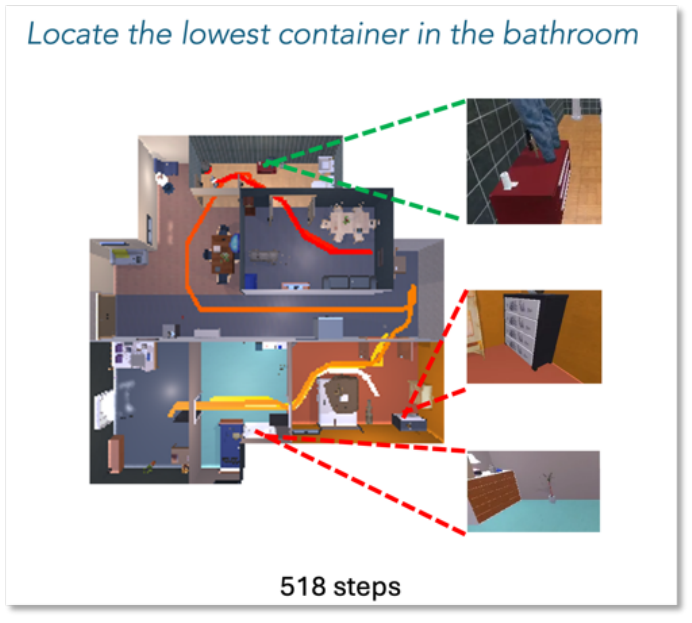}
        \else
        \includegraphics[width=\linewidth]{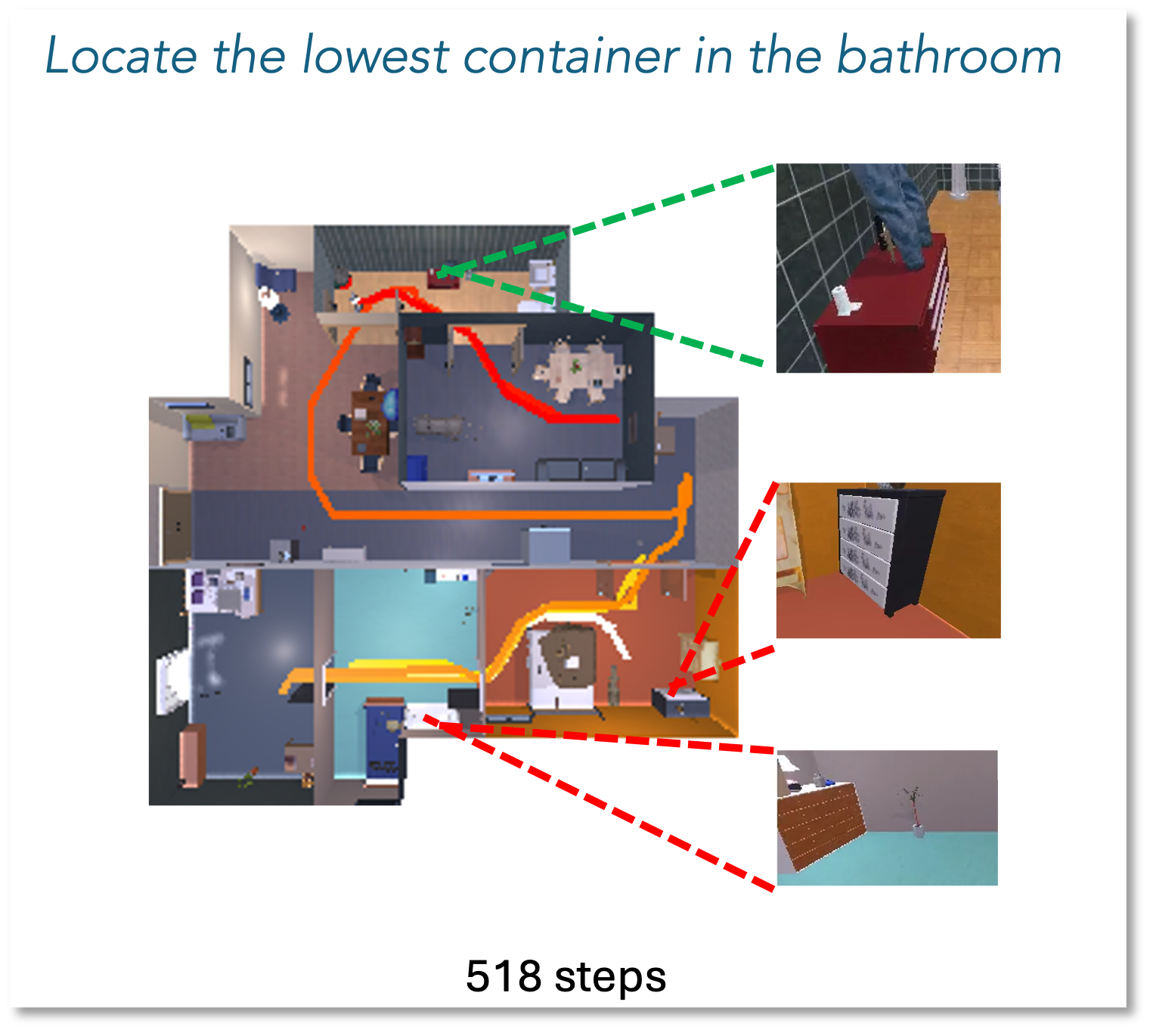}
        \fi
        \caption{ObjNavRelAttr Task}
    \end{subfigure}
    \hfill
    \begin{subfigure}[b]{0.237\textwidth}
                \ifx\arxiv\undefined
        \includegraphics[width=\linewidth]{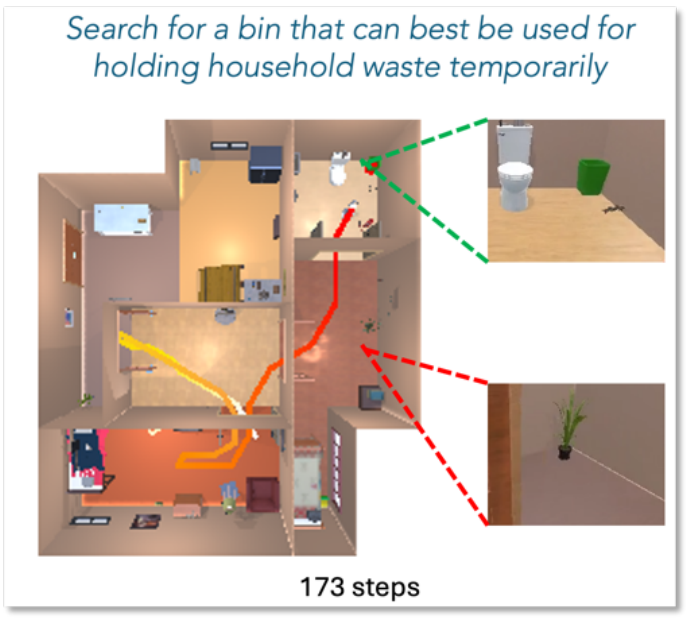}
        \else
        \includegraphics[width=\linewidth]{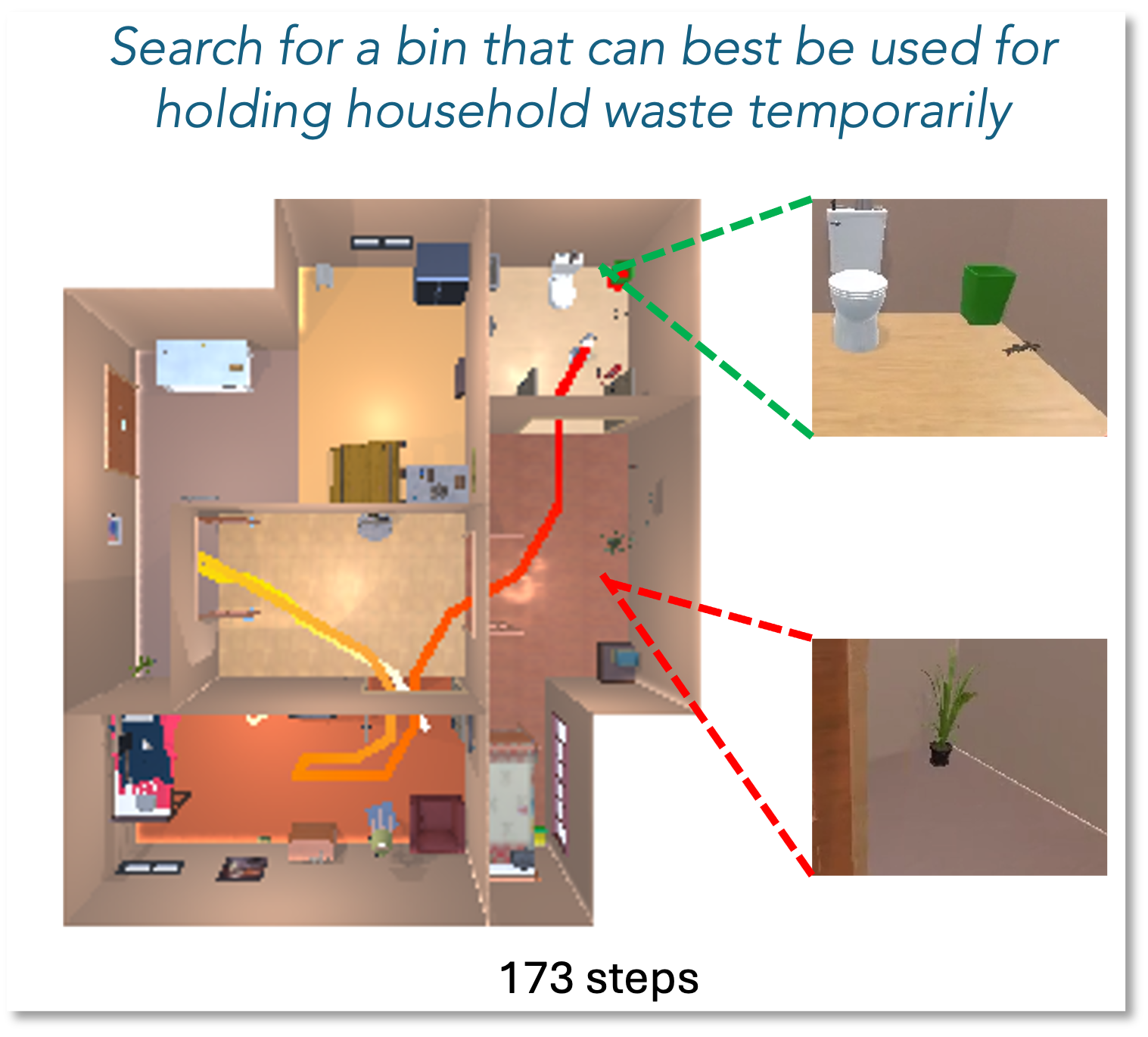}
        \fi
        \caption{ObjNavAfford Task}
    \end{subfigure}
    \hfill
\caption{
We evaluate \methodname{} on mobile manipulation tasks. (a, b) In-distribution tasks, in unseen environments. (c, d) Novel tasks that require unseen capabilities from pretraining, in unseen environments. \methodname{} excels in long-horizon tasks, showing strong object recognition, relational reasoning, and exploration abilities.
}
\label{fig:result_vis}
\vspace{-1em}
\end{figure*}

\section{Results}
We evaluate \methodname{} on a set of navigation and manipulation tasks both in simulation and in the real world. Through our experiments, we seek to answer the following questions: 
\textbf{Q1}: Can \methodname{} achieve state-of-the-art performance on tasks both within and outside the training data of the pre-trained policy?
\textbf{Q2}: Can \methodname{} learn new capabilities never seen during pre-training and generalize to unseen tasks? 
\textbf{Q3}: Can the policies learned by \methodname{} transfer to the real world? 
\textbf{Q4}: Can \methodname{} enable efficient adaptation to new robot embodiments and new behavior?
\textbf{Q5}: Are the stabilization techniques in \methodname{} necessary to ensure stable fine-tuning?

All of the experiments use the same hyperparameters, specified in App.~\ref{app:hyperparameters}. Unless stated otherwise, results for \methodname{} are obtained using sparse rewards that correspond to task completion. 
Visualizations of the robot trajectories are shown in Fig.~\ref{fig:result_vis} and on our project website.

\subsection{\methodname{} on seen capabilities}
\label{ss:indes}

First, we evaluate the performance of \methodname{} in comparison to prior behavior cloning (BC) and reinforcement learning (RL) baselines. Specifically, we test \methodname{} on CHORES-$\mathbb{S}$~\cite{ehsani2024spoc}, a recently introduced simulation benchmark designed for household robot tasks. CHORES-$\mathbb{S}$ encompasses four task types that require various skills, including navigation, object recognition, object manipulation, and environment exploration. Similar to \cite{ehsani2024spoc,zeng2024poliformer}, the policies use the agent's RGB observations as input to predict discrete actions, which represent movements of the base, arm, gripper, and an \textit{END} action to signify task completion. For further details on the action space, observation space, and task definitions, please refer to App.~\ref{app:action}, ~\ref{app:obs}, ~\ref{app:task}.

CHORES tasks are very challenging due to their long-horizon nature, partial observability, RGB-only observation space, and diverse scenes and objects. Therefore, previous methods struggle to complete these tasks.
Since CHORES tasks are contained in the training data of the SPOC model that \methodname{} fine-tunes upon, our goal is to utilize \methodname{} to improve performance on these in-distribution capabilities.

\textbf{Baselines.} Our baselines consist of prior works in imitation learning, reinforcement learning from scratch, and reinforcement learning fine-tuning from pre-trained policies. Aside from \textbf{SPOC}\cite{ehsani2024spoc}, the robot foundation model that we fine-tune upon, we additionally compare against \textbf{Poliformer} \cite{zeng2024poliformer}, a transformer-based RL-from-scratch method that achieved SOTA performance on object navigation; \textbf{EmbSigLIP} \cite{khandelwal2022simple}, a GRU-based RL-from-scratch method; \textbf{PIRLNav} \cite{ramrakhya2023pirlnav}, an RL fine-tuning method that employs learning rate scheduling to warm-start the value function; and \textbf{JSRL} \cite{uchendu2023jump}, an off-policy RL fine-tuning method that gradually “roll in” experiences with the prior policy. 

We compare against baselines in two settings: (1) a fair-comparison setting, where the baseline methods use the same sparse reward as \methodname{}, and (2) an unfair-comparison setting, where the baseline methods use a privileged, task-specific dense reward that has been hand-coded by human experts. It is important to note that each new task demands significant researcher effort to design and curate a dense reward function that avoids collapsing during training and is not scalable to new tasks.

To demonstrate the superiority of \methodname{}, all baseline methods are trained for more steps than \methodname{}. Specifically, the fair-comparison baselines are trained for 3$x$ more steps on ObjectNav and RoomVisit, and 2$x$ more steps on Fetch and Pickup. 
The unfair-comparison baselines are trained until convergence to obtain the best possible result. Notice that this often means significantly longer training time. For example, for the Poliformer (Dense) on ObjectNav, the result is obtained after training for 300M steps - over 15$x$ as many training steps that \methodname{} uses on ObjectNav.

Results are shown in Table.~\ref{tab:indist} and Fig.~\ref{fig:result_vis} (a, b), where we evaluate on \textbf{unseen} simulated houses and report Success rate as well as Episode-length weighted Success (SEL \cite{eftekhar2023selective}) which measures the efficiency of the policies. As shown by the table, \methodname{} not only significantly outperforms the fair-comparison baselines, but outperforms the unfair baseline on three out of the four tasks despite using significantly fewer training steps (\textbf{Q1}). Please find training curves in Fig.~\ref{fig:abla}(a).

\subsection{\methodname{} on novel capabilities}

\label{ss:ood}

A well-trained robotics foundation model should learn features useful for all robotics tasks, not only applicable to in-distribution tasks appearing in its original training data.
To investigate if \methodname{} can take advantage of these pre-trained features, we examine the performance of \methodname{} on a set of novel capabilities never seen by the foundation model.
Specifically, we evaluate \methodname{} on three navigation tasks that specify target objects/locations in different ways and require distinct types of explorations and skills, including 1) ObjNavRelAttr, which identifies target objects through relative object attributes comparison (e.g. ``find the largest apple''); 2) RoomNav, which requires the robot to navigate to room types instead of objects (e.g. ``go to the kitchen''); and 3) ObjNavAfford, which requires object affordance understanding (e.g. ``find something I can sit on''). 
Note that new reasoning skills are required for these unseen tasks; for example, in ObjNavRelAttr, the agent must search the environment for all objects of the specified type, reason about their properties, and issue a completion action when it identifies the correct instance.

We compare against the Poliformer~\cite{zeng2024poliformer} baseline described in Sec.~\ref{ss:indes}, as well as SPOC++, a BC baseline that has the same network architecture as SPOC but uses additional expert demonstrations (1M frames per aforementioned task). Note that these demonstrations are not available to \methodname{}, nor to the SPOC model that \methodname{} fine-tunes.

We show the results in Table~\ref{tab:ood} and Fig.~\ref{fig:result_vis} (c, d). 
On these out-of-distribution tasks that require novel capabilities, \methodname{} achieves state-of-the-art performance without any additional hyperparameter tuning (\textbf{Q2}), even where the baselines have unfair advantages.
It is worth noting that, since specifying each of these new tasks $T_n$ is as simple as specifying a success criteria $R_n$ and the associated language instructions $L_n$, these results imply that we can apply \methodname{} to on-the-fly tasks without much engineering effort. This suggests a path towards continual adaptation.

\begin{table}[t]
\centering
\scriptsize
\setlength{\tabcolsep}{4pt}
\caption{\methodname{} can fine-tune for tasks that are never seen by the base model, and achieve state-of-the-art performance. Baselines with privileged information are \colorbox{lightblue}{\textcolor{black}{\emph{marked in blue}}}. }
\begin{tabular}{c||cc>{\columncolor{lightblue}}c>{\columncolor{lightblue}}c} 
\toprule
                        Success (SEL)& \methodname{} (ours)    & Poliformer (Sp)                            & SPOC++                    & Poliformer (De)                         \\
 \midrule
 
 ObjNavRelAttr       
 & \textbf{71.0 (63.6)}  &  6.7 (6.7)  & 54.5 (44.6)
 &  36.1 (32.4)   
 \\ 
 RoomNav      
 
 & \textbf{91.6 (85.6)}    &   57.0 (51.8)     & 74.5 (59.9)
 &    75.0 (62.4)    
 \\ 
 
 ObjNavAfford      
 
 & \textbf{79.7 (70.6)}  &    35.5 (29.4)     & 62.4 (50.6)
 &    53.8 (43.1)      
 \\ 

 \bottomrule
\end{tabular}
\label{tab:ood}
\vspace{-1em}
\end{table}

\subsection{\methodname{} on real robots}

\begin{figure}[t]
\centering
\includegraphics[width=0.9\linewidth]{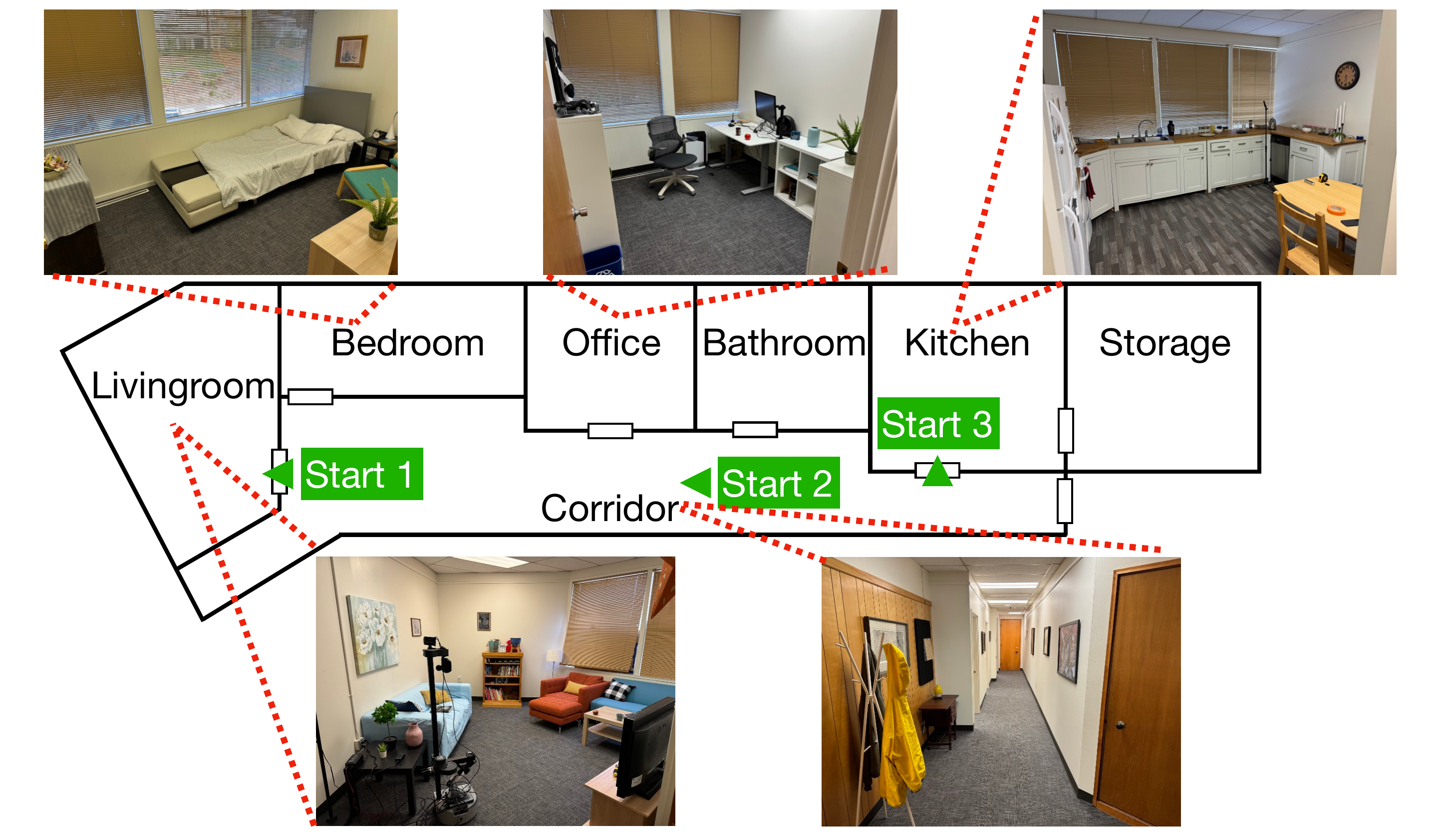}
\caption{The real-world layout that we tested upon}
\label{fig:rw}
\vspace{-2em}
\end{figure}

To examine the performance of \methodname{} on real robots, we evaluate the policies learned by \methodname{} in a real-world apartment on a Stretch RE-1~\cite{kemp2022design}. This layout (Fig.~\ref{fig:rw}) is never seen by the robot during training. 
We directly deploy policies without any adaptation or real-world fine-tuning, and leverage a heuristic object grasping model following SPOC \cite{ehsani2024spoc}.
We compare against SPOC and Poliformer\footnote{Poliformer reported real-world results only for the ObjectNav Task.} with dense reward, and report the results in Table~\ref{tab:real}.
Sim-to-real approaches introduced in Sec.~\ref{ss:sim} enable the successes of \methodname{} in simulation to directly transfer to the real world, achieving state-of-the-art performances on a set of real world navigation and mobile manipulation tasks (\textbf{Q3}).


\begin{table}[t]
\centering
\footnotesize
\caption{Real-world results (total of 46 tasks). For manipulation tasks, we report both full success (policy and heuristic grasping) and policy success (proximity) following \cite{ehsani2024spoc}.}
\begin{tabular}{c||cc>{\columncolor{lightblue}}c} 
\toprule
                        Success Rate& \methodname{} (ours)                   & SPOC                    & Poliformer (Dense)       \\
 \midrule
 
 ObjectNav       
 & \textbf{94.4}    & 50.0
 &   83.3 
 \\ 
 Fetch      
 
 & \textbf{66.7} (55.6)       & 33.3 (11.1)
 &  X         
 \\ 
 
 PickUp      
 
 & \textbf{86.7} (66.7)        & 66.7 (46.7)
 &  X         
 \\ 
 RoomVisit     
 & \textbf{75.0}   & 50.0
 & X   \\ 

 \bottomrule
\end{tabular}
\label{tab:real}
\end{table}

\subsection{\methodname{} for adaptation}

\methodname{} opens up the possibilities for learning behaviors not captured by the demonstration data (and thus the foundation robotics model). We examine this in two setups, cross-objective and cross-embodiment capabilities of \methodname{} (\textbf{Q4)}.

\subsubsection{Adaptation to New Embodiment}
We use \methodname{} to fine-tune SPOC (which is trained only on Stretch-RE1) to adapt to Locobot \cite{gupta2018robot}.
Locobot has different action space and camera parameters: it lacks the manipulation degrees-of-freedom that Stretch possesses, but has a rotatable, narrower field-of-view camera mounted lower. To facilitate cross-embodiment transition, we simply mask out the invalid actions output by the policy, and repurpose two of the invalid actions to control the camera. \methodname{} effectively utilizes the pre-trained policy to adapt to the new embodiment, as shown by the table below on the ObjectNav Task:

\begin{table}[h]
\vspace{-10pt}
\centering
\footnotesize
\begin{tabular}{c||cc} 
\toprule
     New Embodiment                   & Success Rate $\uparrow$            & SEL  $\uparrow$     \\
 \midrule
 
 \methodname{}       
 & \textbf{72.0}    & \textbf{47.2}
 \\ 
 Poliformer zero-shot\tablefootnote{We zero-shot evaluate Poliformer~\cite{zeng2024poliformer} $400$M ckpt trained with LoCoBot. This baseline was trained in ProcTHOR-10k, instead of ObjaTHOR houses.}      
 
 & 57.5        & 30.1
 \\ 
 Poliformer (Sparse Reward)     
 
 & 44.0        & 29.7
 \\

 \bottomrule
\end{tabular}
\vspace{-1em}
\end{table}

\subsubsection{Adaptation to New Behavior}
We investigate whether \methodname{} can be used to shape a robot's behavior after the policy is trained, using only a few training steps. We test two new behaviors: 1) encouraging the agent to be more efficient (+step penalty $-0.01$/step), and 2) reducing the number of unwanted collisions with the environment (+collision penalty $-0.5$/collision). By adding a reward term tailored to each behavior, the policy adapts to these new behaviors after just 6 hours of training, with minimal impact on the success rate. The following table presents the results for the Fetch task:

\begin{figure}[t]

    \begin{subfigure}
    {0.24\textwidth}
\centering\includegraphics[trim=0 20 0 20, clip, width=0.99\linewidth]{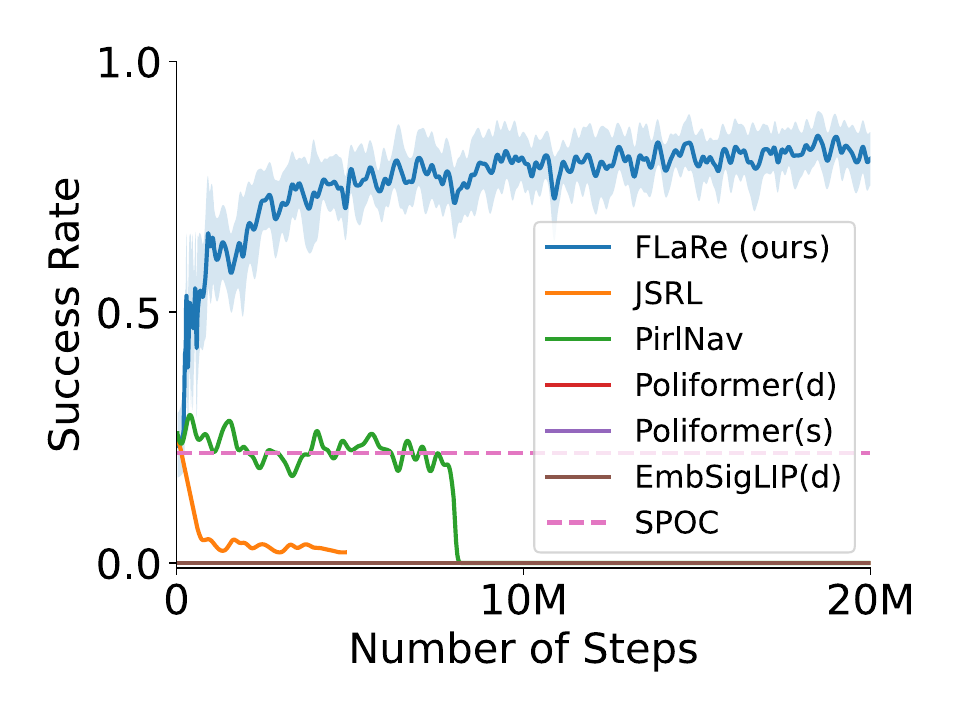}
        \caption{Baselines}
    \end{subfigure}
    \hfill
    \begin{subfigure}{0.24\textwidth}
        \centering\includegraphics[trim=0 20 0 20, clip, width=0.99\linewidth]{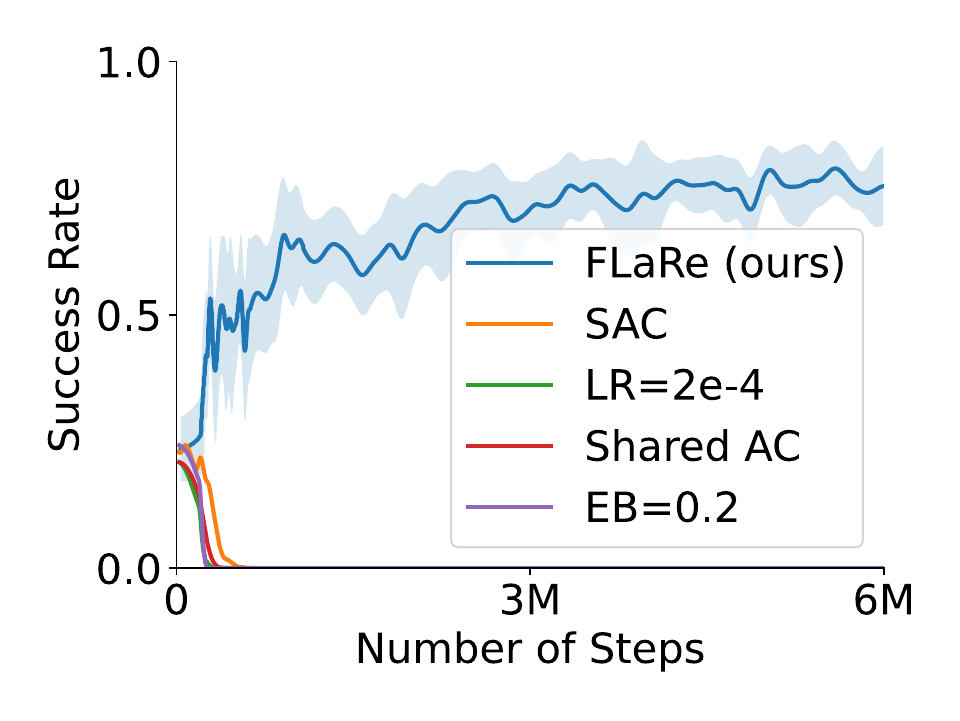}
        \caption{Ablations}
    \end{subfigure}

\caption{Baseline performances and ablation studies on the Fetch task. \methodname{} is the only method that can achieve good performance on this challenging task.}
\label{fig:abla}
\end{figure}

\begin{table}[h]
\vspace{-8pt}
\centering
\scriptsize
\setlength{\tabcolsep}{4pt}
\begin{tabular}{c||ccc} 
%
%
 \toprule
      New Behaviors                  & Success Rate $\uparrow$            & Episode Length  $\downarrow$                  & \# of Collisions   $\downarrow$     \\
 \midrule
 
 \methodname{}       
 & \textbf{66.9}    & 258.2
 &   10.0
\\ 
 + Step Pen.        
 
 & 65.7        & \textbf{222.8}
 &  10.0 
 \\ 
 + Coll. Pen.        
 
 & 66.7        & 251.2
 &  \textbf{3.1}         
 \\

 \bottomrule
\end{tabular}
\vspace{-2em}
\end{table}


\subsection{Ablation studies}
\label{ss:abla}
To evaluate whether the techniques proposed in Sec.~\ref{ss:ft} are necessary for the performance of \methodname{}, we evaluate four ablation variants of our method. To evaluate whether using on-policy methods is important, we tested switching the PPO algorithm by Soft Actor-Critic~\cite{haarnoja2018soft} (\textbf{SAC}). To evaluate whether a small learning rate is necessary, we tested setting the learning rate to $2e-4$, 10 times our original learning rate. To evaluate the importance of having separated actor and critic, we tested \textbf{Shared AC}, where the actor and critic share the transformer encoder and decoder trunk. Finally, we tested \textbf{EB=0.2}, which set the coefficient of the entropy bonus in PPO to 0.2.
We show the training curves in Fig.~\ref{fig:abla}(b).

Perhaps surprisingly, removing any single one of the stabilizing techniques in \methodname{} results in the success rate quickly collapsing to 0 on the fetch task, while \methodname{} learns very robustly with the same set of hyperparameters across variety of tasks and experiment setups (\textbf{Q5}). This showcases the importance of all of the techniques introduced in \methodname{}.

\section{CONCLUSIONS}

\methodname{} is an efficient and scalable approach for fine-tuning large-scale robot policies using RL. It enables effective adaptation to unseen tasks and achieves state-of-the-art performance in both simulation and real-world settings. \methodname{}'s adaptability to new embodiments and behaviors unlocks the potential for flexible deployment across a wide range of robotic platforms.
\methodname{}'s main limitation lies in its reliance on simulation environments for fine-tuning.
While leveraging recent work in simulation generation~\cite{deitke2023phone2proc, torne2024reconciling} offers a promising direction, tackling tasks where robust simulations are unavailable—such as those involving liquids or soft objects—remains challenging and may require fine-tuning directly in the real world.


\clearpage

\addtolength{\textheight}{-6cm}   






\bibliographystyle{IEEEtran}
\bibliography{references}

\ifx\arxiv\undefined
\else
\addtolength{\textheight}{6cm}   


\section*{\Large APPENDIX}

\subsection{Results Visualization}
\label{app:visualization}
We encourage the reader to visit our website (\url{robot-flare.github.io}) for visualizations of trajectories generated by \methodname{} both in simulation and in the real world, including performances visualization, behavior analysis, and failure mode analysis.

\subsection{Hyperparameter}

\begin{table}[h!]
    \centering
    \begin{tabular}{l l}
        \toprule
        \multicolumn{2}{c}{\textbf{Training and Model Details}} \\
        \midrule
        \textbf{Parameter} & \textbf{Value} \\
        \midrule
        Total Rollouts & 32 \\
        Learning Rate & 0.0002 \\
        Mini Batch per Update & 1 \\
        Update Repeats & 4 \\
        Max Gradient Norm & 0.5 \\
        Discount Value Factor $\gamma$ & 0.99 \\
        GAE $\lambda$ & 0.95 \\
        PPO Surrogate Objective Clipping  & 0.1 \\
        Value Loss Weight & 0.5 \\
        Entropy Loss Weight & 0.0 \\
        Steps for PPO Update & 128 \\
        Transformer State Encoder Layers & 3 \\
        Transformer State Encoder Hidden Dims & 512 \\
        Transformer State Encoder Heads & 8 \\
        Causal Transformer Deocder Layers & 3 \\
        Causal Transformer Deocder Hidden Dims & 512 \\
        Causal Transformer Deocder Heads & 8 \\
        \bottomrule
    \end{tabular}
    \vspace{1mm}
    \caption{Hyperparameters for training and model architecture. We use AllenAct~\cite{weihs2020allenact} to implement our models and conduct experiments.
    }
    \label{tab:hyperparams}
\end{table}

\subsection{Number of Training Steps}
\label{app:hyperparameters}
The base SPOC model that we evaluted and fine-tuned upon is trained for 50k gradient update steps on a total of 100k episodes of demonstrations across the CHORES tasks, where the training hyperparameter and training data is exactly the same as in the original SPOC paper.

For navigation tasks that do not involve manipulating objects (i.e. ObjectNav and RoomVisit), \methodname{} performs RL fine-tuning for 20M steps, while all other fair-comparison baseline methods perform RL training for 60M steps. 
For mobile manipulation tasks (i.e. Fetch and Pickup), \methodname{} performs RL fine-tuning for 50M steps, while all other fair-comparison baseline methods perform RL training for 100M steps.
For adaptation tasks, we run \methodname{} fine-tuning for 50M steps on ObjNavRelAttr and ObjNavAfford, and 20M steps on RoomNav. For cross-embodiment, we run \methodname{} for 30M steps.

All of the aforementioned experiments use the same base SPOC mode, with exactly the same set of hyperparameters.

\subsection{CHORES Benchmark}
\label{app:benchmark}
A big portion of \methodname{}'s evaluation is carried out on the CHORES benchmark. We provided detailed information about this benchmark, including the observation space, action space, and task specifications.

\subsubsection{Observation Space}
\label{app:obs}
The observation space of CHORES consists of two ego-centric 384 $\times$ 224 RGB camera pointing towards orthogonal directions, where one points towards the direction of navigation and the other points at the arm. Additionally, a natural language text instruction is re-sampled at the start of each episode and appended to the observation to specify what the robot should be doing.

\subsubsection{Action Space}
\label{app:action}
The action space of CHORES consists of 20 discrete actions: Move Base ($\pm$ 20 cm), Rotate Base ($\pm$6$^{\circ}$,
$\pm$30$^{\circ}$), Move Arm (x, z) ($\pm$2 cm, $\pm$10 cm), Rotate Grasper
($\pm$10$^{\circ}$), pickup, dropoff, done with subtask, and terminate.

\subsubsection{Tasks Specifications}
\label{app:task}
We describe the CHORES tasks in Table.~\ref{tab:bench_tasks}. For each task, if the robot exceeds the maximum steps, the episode is terminated and marked as failed.

\begin{table*}[t]
    \centering
    \footnotesize
    \resizebox{1.5\columnwidth}{!}{%
        \begin{tabular}{|p{0.15\columnwidth}|p{0.9\columnwidth}|c|}
            \hline
            \textbf{Task} & \textbf{Description \& Example} & \textbf{Max Steps} \\
            \hline
            ObjectNav & Locate an object category: ``find a mug'' & 600 \\
            \hline
            PickUp & Pick up a specified object in agent line of sight: ``pick up a mug'' & 600\\
            \hline
            Fetch & Find and pick up an object: ``locate a mug and pick up that mug'' & 600 \\
            \hline
            RoomVisit & Traverse the house. ``Visit every room in this 5-room house. Indicate when you have seen a new room and when you are done.'' & 1000\\
            \hline
        \end{tabular}%
    }
    \vspace{-0.1cm}
    \caption{CHORES tasks.}
    \label{tab:bench_tasks}
    \vspace{-1em}
\end{table*}

For each task, we splited a total of 191,568 houses from ProcThor\cite{deitke2022️} into training and testing sets with a ratio of 10:1, to ensure that the test evaluation is conducted in unseen houses.


\begin{figure*}[t]
\centering
\includegraphics[width=0.9\textwidth]{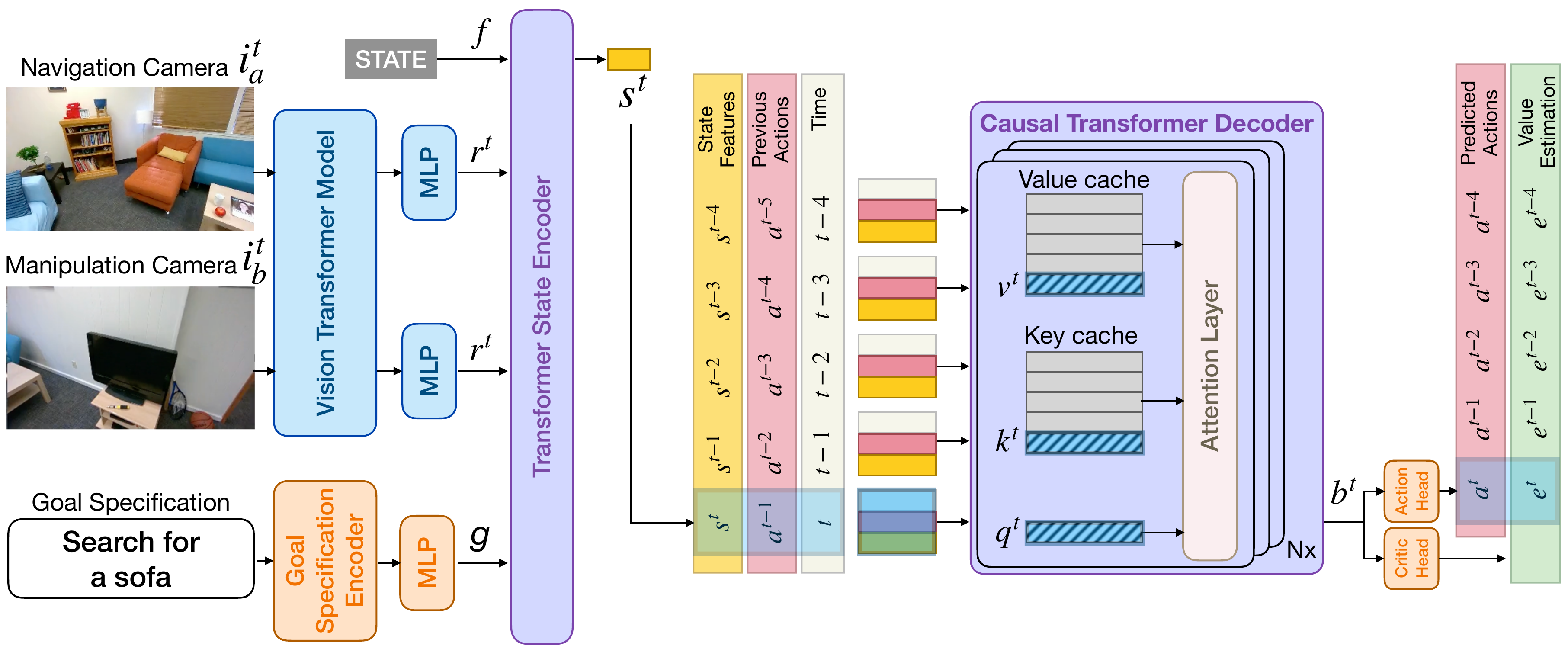}
\caption{ A visualization of the network architecture of the transformer-based SPOC model that \methodname{} fine-tunes upon. }
\label{fig:model_details}
\vspace{-1em}
\end{figure*}

\subsection{The SPOC Model}
\label{app:spoc_model}
In this work, we use a slightly modified version of the SPOC model~\cite{ehsani2024spoc} inspired by Poliformer~\cite{zeng2024poliformer}, where the transformer decoder block in SPOC is replaced by the decoder from Llama 2 LLM \cite{touvron2023llama} to speed up training and inference. 
At each step, the SPOC model takes in the new observations consisting of two RGB images and a text instruction. Each of these images are separately passed through a frozen \textbf{vision transformer model} (DinoV2\cite{oquab2023dinov2}) to extract a set of visual tokens. These tokens, along with an embedding of the natural language instructions using a pre-train text encoder T5\cite{ni2021sentence}, are summarized by a \textbf{transformer state encoder} to produce the observation representation. A \textbf{causal transformer decoder} then decodes the observations feature across all steps within the current episode into a belief vector that is passed through an actor head to generate the action prediction. We provide a visualization of our model in Fig.~\ref{fig:model_details}, and explain each of these components in detail below.

\subsubsection{Vision Transformer Model}
We use DINOv2 as the visual foundation backbone because of its remarkable ability to make dense predictions that generalize across sim and real.
Our input to the visual backbone are two RGB observations $i_a$ and $i_b$. $i_a \in \mathbb{R}^{H \times W \times 3}$ is captured by the navigation camera and $i_b \in \mathbb{R}^{H \times W \times 3}$ is captured by manipulation camera, where $H$ and $W$ are the height and width of the image. The visual backbone then produces a patch-wise representation $r \in \mathbb{R}^{\frac{H}{14} \times \frac{W}{14} \times h}$, where $h$ is the hidden dimensions of the visual representations.
$r$ is then reshaped and projected to generate visual tokens $v_\text{raw} \in \mathbb{R}^{n_\text{patch} \times d_\text{encoder}}$. A learnable camera-type embedding is then added to this visual tokens to ensure the model can differentiate between the navigation and the manipulation cameras, resulting in the final visual features $v$. 
To ensure sim-to-real transfer, we freeze the DinoV2 weight throughout training.

\subsubsection{Transformer State Encoder}
This module summarizes the observations at each timestep as a vector $s \in \mathbb{R}^d$.
The input to this encoder includes the visual representation $v$, the text feature $g$, and a learnable STATE token $f$. We concatenate these features together and feed them to a non-causal transformer encoder. This encoder then returns the output corresponding to the STATE token as the state feature vector. The transformer state encoder digests both visual and text features, and can thus be seen as generating a text-conditioned visual state representation.

\subsubsection{Causal Transformer Decoder} To deal with partial observability and handle long-horizon tasks, SPOC uses a causal transformer decoder to perform explicit memory modeling over time. The causal transformer decoder consumes the visual representations generated by the transformer state encoder, additively combines them with sinusoidal temporal position encodings and learned previous time step action embeddings, and generates the belief vector used for action generation.

\subsection{Real Robot Setup}
\label{app:real_world}
Following SPOC~\cite{ehsani2024spoc}, we equipped our Stretch RE-1 robot with two identical Intel
RealSense 455 fixed cameras, namely the navigation and
the manipulation camera. These cameras have a vertical field of view of 59$^\circ$ and are capable of capturing 1280×720 RGB-D images. Both of these cameras point slightly down, with the horizon at a nominal 30$^\circ$, to optimize the agent’s perspective of its functional workspace.
The images returned by these cameras are first resized to 396 $\times$ 224, and the cropped to 384 $\times$ 224, to match the image observations during training.

Same as SPOC, we assess the performance of our models on ObjectNav and Fetch in a 6-room apartment also used in Phone2Proc~\cite{deitke2023phone2proc},
Pickup in RoboThor~\cite{deitke2020robothor}, and RoomVisit in both environments. The 6-room apartment contains environment variations wholly unseen at train time, including a new configuration (multiple rooms off a long corridor), two new room
types (office and corridor), rooms with non-orthogonal wall alignment, and many unseen object instances. For each object in ObjectNav and Fetch, we tested three starting positions: once from the living room, once from the middle of the corridor, and once from the kitchen. We visualize these starting locations in Fig.~\ref{fig:rw}. Below, we provide objects that we tested upon in the real world for each tasks.

\subsubsection{ObjectNav} Target objects are Sofa, Bed, Chair, Apple,
Vase, and Houseplant, each from three starting positions.

\subsubsection{Fetch} Target objects are Apple, Vase, and Houseplant
from the same three starting positions. In one small change
from ObjectNav episodes, object instances are replaced with
instances which better fit into Stretch’s grasping envelope
and in some cases at a better height for interaction, but availability and placement are nearly identical.

\subsubsection{PickUp}
Objects are placed on three different surfaces
(coffee table, desk, and nightstand) at three different
heights. Objects are Apple, Houseplant, Spray Bottle, Mug,
and Vase. 

\subsubsection{RoomVisit}
The full 6-room apartment is explored, and
then partitioned into two 3-room apartments to evaluate the
ability of SPOC to explore large and small spaces. We additionally explore a section of RoboTHOR and attached workroom as a novel 3-room apartment.
\fi

\end{document}